\documentclass[letterpaper]{article}
\usepackage{amsfonts} 
\usepackage{amssymb} 
\usepackage{amsmath} 
\usepackage{graphicx} 
\usepackage{subfigure} 
\usepackage{array} 
\usepackage{hyperref} 
\usepackage{authblk} 
\usepackage{bbold}
\usepackage[letterpaper, margin=1.0in]{geometry}

\graphicspath{{./figs/}}

\usepackage{amsmath}

\DeclareMathOperator{\tr}{tr}

\begin{document}


\title{Estimation of matrix trace using machine learning}
\author{Boram Yoon}
\affil{Los Alamos National Laboratory, CCS-7, Los Alamos, New Mexico 87545}

%
%
\date{}

\maketitle


\begin{abstract}
We present a new trace estimator of the matrix whose explicit form is not given but its matrix multiplication to a vector is available. The form of the estimator is similar to the Hutchison stochastic trace estimator, but instead of the random noise vectors in Hutchison estimator, we use small number of probing vectors determined by machine learning. Evaluation of the quality of estimates and bias correction are discussed. An unbiased estimator is proposed for the calculation of the expectation value of a function of traces. In the numerical experiments with random matrices, it is shown that the precision of trace estimates with $\mathcal{O}(10)$ probing vectors determined by the machine learning is similar to that with $\mathcal{O}(10000)$ random noise vectors.
\end{abstract}
\vspace{2cm}

%
%
%
%
\section{Introduction}
\label{sec:into}
Many applications require the trace calculation of the matrix whose explicit form is not given but only its matrix multiplication to an arbitrary vector is available. An example is the trace calculation of the inverse of Dirac matrix in lattice quantum chromodynamics (QCD) \cite{Bhattacharya:2015esa}. The Dirac matrix is large and sparse so that explicit matrix inversion is not feasible, but the matrix multiplication of the inverse matrix to a vector is available through iterative linear equation solvers, such as the Conjugate Gradient (CG) algorithm \cite{Hestenes&Stiefel:1952}. In general, an exact solution can be obtained by $N$ matrix-vector multiplications for a $N\times N$ matrix. However, such an exact estimation is computationally expensive for large matrices, so stochastic approaches, such as the Hutchison estimator\cite{Hutchinson89}, are widely used. For specific structures of matrices, such as the banded matrices, improved trace estimators can be defined by exploiting the matrix structure, so that precise trace estimates are obtained by small number of matrix-vector multiplications \cite{Bekas:2007,Tang:2012,Stathopoulos:2013aci}. In general, however, the matrix structure is too complicated to be exploited by human knowledge, while it can be explored by using machine learning.

Machine learning is a field of science that makes a computer to act based on the learning from data. In contrast to the specific models with small number of free parameters in conventional data analysis, machine learning uses general models with large number of free parameters, such as the artificial neural network (ANN) \cite{McCulloch1943, Werbos74} or the support vector machines (SVMs) \cite{Boser:1992:TAO:130385.130401}. Based on the generality of the learning model and large number of free parameters, computer builds its own model from data. 

In this paper, the idea of machine learning is applied to the trace estimation. We build a trace estimator with high degrees of freedom, and train the estimator by using large number of matrices that have similar structures. As other machine learning applications, the learning procedure is computationally expensive, but the trace estimation using the trained estimator requires only small number of matrix-vector multiplications. In section~\ref{sec:theory}, we define the new trace estimator and learning procedure, and discuss the quality of the estimates and bias correction. Numerical experiments of the new trace estimator are shown in Section~\ref{sec:expt}, and we conclude in Section~\ref{sec:sum}.

\section{Theory}
\label{sec:theory}
\subsection{Backgrounds}
Let us consider a $L\times L$ matrix $A$ where the matrix itself is difficult to obtain, but its product to a vector is easy to evaluate. The exact trace of the matrix can be calculated by 
\begin{align}
\tr(A) = \sum_{i=1}^L \mathbf{e}^\mathsf{T}_i A \mathbf{e}_i
\label{eq:tr_exact}
\end{align}
where $\mathbf{e}_i$ is the vector whose $i$-th component is $1$ and other components are $0$. This method requires $L$ times of matrix-vector multiplications, so it is computationally very expensive for large matrices.

The trace of the matrix can be estimated by using the stochastic estimator
\begin{align}
  \tr(A) \approx \frac{1}{N_z} \sum_{i=1}^{N_z} \mathbf{z}^\mathsf{T}_i A \mathbf{z}_i
  \label{eq:tr_stoch}
\end{align}
with the random noise vectors $\mathbf{z}_i$ satisfying
\begin{align}
  \frac{1}{N_z}\sum_{i=1}^{N_z} \mathbf{z}_i 
    = \mathcal{O}\left(\frac{1}{\sqrt{N_z}}\right), \qquad
  \frac{1}{N_z}\sum_{i=1}^{N_z} \mathbf{z}_i \mathbf{z}^\mathsf{T}_i 
    = \mathbb{1} + \mathcal{O}\left(\frac{1}{\sqrt{N_z}}\right).
\end{align}
The uncertainty of the stochastic trace estimator depends on the structure of the matrix $A$ and type of the random noise $\mathbf{z}$, because some type of random noise cancel the contribution from the elements of the matrix that cannot contribute to the trace but only to noise for the matrices that have specific structure \cite{Bhattacharya:2015gma}. One popular choice of the random noise vector is to filling the vector elements following Rademacher distribution\footnote{$1$ or $-1$ with probability $1/2$.}. Corresponding trace estimator is called the Hutchinson estimator \cite{Hutchinson89}.

When structure of the matrix $A$ is known, one could use a small number of sophisticatedly chosen probing vectors, instead of the large number of random noise vectors, for the precise trace estimation \cite{Coleman1983EoS,Bekas:2007,Tang:2012}. It is called the probing method. It also can be applied to a trace estimation for the inverse of a sparse matrix $M^{-1}$ by approximating the structure of $M^{-1}$ using the sparsity pattern of the powers of the matrix, $M^{n}$ \cite{Tang:2012,Stathopoulos:2013aci}. When structure of the matrix is unknown or too complex to be exploited, however, one cannot construct such probing vectors by hand. If there is a large set of matrices that have similar structures, one may consider to find the probing vectors by using those data set instead of constructing the probing vectors by using human knowledge.

\subsection{Construction of Probing Vectors using Machine Learning}
Our goal is to find a set of probing vectors $\{\mathbf{p}_{l}\}$ for $l=1,2,3,\ldots,N_p$ that probes the trace of a matrix $A$ as
\begin{align}
  \tr(A) \approx \sum_{l=1}^{N_p} \mathbf{p}^\mathsf{T}_l A \mathbf{p}_l\,,
  \label{eq:probe_est}
\end{align}
for a given number of probing vectors $N_p$, by using the machine learning. The machine learning requires input data to train the probing vectors: a set of matrices that have similar structures, and the traces of those matrices. Let us consider a set of sample matrices $\{M_{t=1,2,\ldots,N}\}$ from a distribution. The cost function, which shows the difference between the exact answer and the probing vector estimate, is defined by
\begin{align}
  Q_t\left(\{\mathbf{p}_l\}\right) = \left(\sum_{l=1}^{N_p} \mathbf{p}^\mathsf{T}_l M_t \mathbf{p}_l - \tr(M_t)\right)^2.
  \label{eq:cost_ftn}
\end{align}
The main idea is to find the probing vectors that minimize the cost function. In this paper, we consider the online machine learning, which updates the probing vectors using only one input training data (matrix) at each update step, as we find that the online learning is more efficient than the batch learning, which uses more than one input training data in an update step, in this application. 

In the practical calculation, the trace of the matrices $\tr(M_t)$ are estimated by using the Hutchison estimator in Eq.~\eqref{eq:tr_stoch} as we are considering matrices whose exact traces are difficult to calculate. In the numerical test given in Section \ref{sec:expt}, we find that a sloppy estimator of $\tr(M_t)$ with relatively small number of random noise vectors can construct the probing vectors that yield precise estimates. In addition to the sloppy estimator, reusing an input matrix and its trace estimate multiple times in the training procedure can further reduce the trace estimation cost. Details are given in Section \ref{sec:expt}.

Let us define a set of probing vectors at a given training time $t$ as $w_t = \{\mathbf{p}_l\}_t$. In the training process, they are updated as follows
\begin{align}
\begin{split}
  \Delta w_t &= \gamma_t \nabla Q_t(w_t) + \eta \Delta w_{t-1}\,, \\
  w_{t+1}    &= w_t - \Delta w_t\,,
\end{split}
  \label{eq:w_update}
\end{align}
for $t = 1,2,3,\cdots$ with initial condition $\Delta w_0=0$. In the numerical experiments, we fill the initial probing vectors $w_1$ by using randomly chosen numbers following normal distribution of $\mathcal{N}(0,1^2)$, but one could use elaborately chosen initial guesses. Here $\nabla Q(w_t)$ is the gradient of the cost function with respect to the elements of the probing vectors, which can be calculated analytically, and $\gamma_t$ is the parameter called the learning rate. With properly chosen learning rate, the update moves the probing vectors towards the steepest descent direction reducing the cost function. In addition to the steepest descent method, we include the momentum term $\eta \Delta w_{t-1}$ to improve the speed of convergence \cite{Rumelhart:1988}. In our numerical experiments, we use $\eta = 0.5\sim0.8$ depending on the other parameters.

Determination of the learning rate $\gamma_t$ is important to make sure that the probing vectors efficiently converge to the global minimum that yields the most precise trace estimate. If it is too small, it converges slowly, and if it is too large, it does not converge to the minimum but oscillate. In early stage of the training, we determine the $\gamma_t$ by using a one-dimensional minimizer in order to maximize the training efficiency. After 100 updates, we calculate $\gamma_0$ by the median of the $\gamma_t$ values of the first 100 updates, and scale the learning rate as
\begin{align}
  \gamma_t = \frac{\gamma_0}{1 + \alpha t} \qquad (\textrm{for } t>100)\,,
  \label{eq:learning_rate}
\end{align}
where $\alpha$ is the parameter that determines the speed of scaling. When $t \lesssim 1/\alpha$, the training actively updates the probing vectors to capture the structure of each input matrix, and when $t \gg 1/\alpha$, the training fine-tunes the probing vectors to describe the general structure of input matrices.

\subsection{Quality of Estimates and Bias Correction}
Once the training is over, the probing vectors can be used for the estimation of trace of a matrix as Eq.~\eqref{eq:probe_est}. Naturally, the estimate may not be exact, and it could yield somewhat different value from the exact trace of the matrix. Considering a set of test matrices $\{M_i\}$, one can define the deviation of the estimate from the exact trace by
\begin{align}
  d_i = \tr(M_i) - \sum_{l=1}^{N_p} \mathbf{p}^\mathsf{T}_l M_i \mathbf{p}_l\,.
  \label{eq:est_err}
\end{align}
The quality of estimates can be assessed based on two factors: (1) non-zero mean value of $d_i$, which gives the bias of estimate and (2) fluctuation (standard deviation) of $d_i$ over different matrices $M_i$, which is related to the accuracy of the estimate.

The bias can be corrected easily by redefining the trace estimator as
\begin{align}
  \tr(A) \approx \widetilde\tr(A) \equiv \sum_{l=1}^{N_p} \mathbf{p}^\mathsf{T}_l A \mathbf{p}_l + \overline{d} \,,
  \label{eq:probe_est2}
\end{align}
where $\overline{d}$ is an average of $d_i$ over a fixed number of sample matrices. In general, one needs to determine the $\overline{d}$ more precise than the fluctuation of $d_i$ so that we can ignore the bias in the error estimation. Since the statistical uncertainty of an average over $N$ samples is $1/\sqrt{N}$ of the standard deviation, averaging over 100 samples of $d_i$ will give a precise enough estimation.

Practically, the exact value of $d_i$ is very difficult to obtain because it requires the exact value of $\tr(M_i)$. Instead of the exact $\tr(M_i)$ calculation, one can use the stochastic estimates of those. Due to the uncertainty of the stochastic estimation, one would need large number of samples of $d_i$ to determine $\overline{d}$ precisely. However, the computation cost can be negligible if one reuses the stochastic trace estimates used in the training procedure.

After the $\overline{d}$ correction, uncertainty of the trace estimator Eq.~\eqref{eq:probe_est2} depends only on the fluctuation of $d_i$. If $d_i$ is distributed normally, one can report the standard deviation of $d_i$ from the set of test matrices $\{M_i\}$ as an uncertainty of the trace estimate. Otherwise, one can devise a proper error estimation based on the distribution of $d_i$.

If one calculates an expectation value of a function of traces for a set of matrices, an unbiased estimator can be defined by applying the technique used in Refs.~\cite{Bali:2009hu, Blum:2012uh} as
\begin{align}
  \Big\langle f\big(\tr(M)\big) \Big\rangle 
    \approx \frac{1}{N}\sum_{i=1}^N f\big(\widetilde\tr(M_i)\big)
    + \frac{1}{N_c}\sum_{j=1}^{N_c} \Big[ f\big(\tr(M_j)\big) - f\big(\widetilde\tr(M_j)\big)\Big]\,,
  \label{eq:unbiased_est}
\end{align}
The first term on the right hand side is the average calculated by using the probing vector trace estimator defined in Eq.~\eqref{eq:probe_est2}. The second term on the right hand side is the bias correction term that takes care of all the misestimation of the trace using the probing vectors. Here we assume that $M_i$ are independent and identically distributed, so use only the first $N_c$ matrices $\{M_1, M_2\, \cdots, M_{N_c}\}$ for the correction term, but they could be chosen randomly or maximally separated in the sequence of $M_i$. The $\tr(M_j)$ in the equation can be any type of unbiased trace estimator, such as the exact estimator in Eq.~\eqref{eq:tr_exact} or the stochastic estimator in Eq.~\eqref{eq:tr_stoch}.

The role of the second term is following. First, let us take expectation value to the right hand side of Eq.~\eqref{eq:unbiased_est}. The terms using the biased estimator $\widetilde\tr(M_i)$ cancel, and the only remaining is the term using unbiased estimator, $\tr(M_j)$. Hence the estimator defined in Eq.~\eqref{eq:unbiased_est} is unbiased. Second, let us take variance to the right hand side of Eq.~\eqref{eq:unbiased_est} in order to see the statistical uncertainty of the estimate. In addition to the variance of the first term, the variance of the second term and the correlation between the first and second term comes in. In this way, the correction term converts the systematic uncertainty of the trace estimator using probing vectors to the statistical uncertainty, keeping the final estimates unbiased. If one uses the unbiased estimator defined in Eq.~\eqref{eq:unbiased_est}, one does not need to include the $\overline{d}$ bias correction term in Eq.~\eqref{eq:probe_est2} because the bias will be corrected automatically in the new definition of the estimator. However, including the $\overline{d}$ correction might reduce the statistical uncertainty in final estimation in some cases.

The size of statistical uncertainty increased by the correction term is depending on two factors: (1) the correlation between $f\big(\tr(M_j)\big)$ and $f\big(\widetilde\tr(M_j)\big)$, and (2) the number of samples used in the correction term, $N_c$. If the trace estimate using probing vectors is good enough so that the function values from those two trace calculations $f\big(\tr(M_j)\big)$ and $f\big(\widetilde\tr(M_j)\big)$ are very close, the contribution of the correction term to the overall statistical uncertainty would be small, and one could use small $N_c$. If the trace estimate using probing vectors is not precise and the two function values are quite different, one would need to use large $N_c$ to suppress the increase of statistical error due to the correction term. In general, we use the unbiased estimator Eq.~\eqref{eq:unbiased_est} when we can take $N_c \ll N$.

\section{Numerical Experiments}
\label{sec:expt}

In this section, we demonstrate the performance of the trace estimator using probing vectors described in the previous section. For the numerical test, we train probing vectors to estimate the trace of inverse of the random $L\times L$ matrix $M$ generated by
\begin{align}
  M(i,j) = \left\{
    \begin{array}{ll}
       1.0            & \textrm{if } i=j,\\
      +0.5 + \xi_{ij} & \textrm{if } i = j + 1 \ (\textrm{mod }L),\\
      -0.5 + \xi_{ij} & \textrm{if } i = j - 1 \ (\textrm{mod }L)
    \end{array} \right.
    \label{eq:rand_mat}
\end{align}
where $\xi_{ij}$ are independent and identically distributed random variables following normal distribution of $\mathcal{N}(0,0.1^2)$. Note that this matrix is an discrete symmetric derivative with identity: $\sum_j M(i,j) f(j) = \frac12\big(f({i-1})-f({i+1})\big)+ f(i)$, except the Gaussian noise. It has the similar form with the Dirac operator in Lattice QCD \cite{degrand2006lattice,gattringer2009quantum}. This is a sparse matrix whose inverse is expensive to calculate, but the matrix-vector product of the inverse matrix can be calculated by using iterative methods. In this numerical test, we use BiCGSTAB algorithm\cite{vanderVorst:1992} to evaluate the product of the inverse matrix $M^{-1}$ and an arbitrary vector.

\begin{figure}[tb]
\centering
  \includegraphics[width=0.49\textwidth]{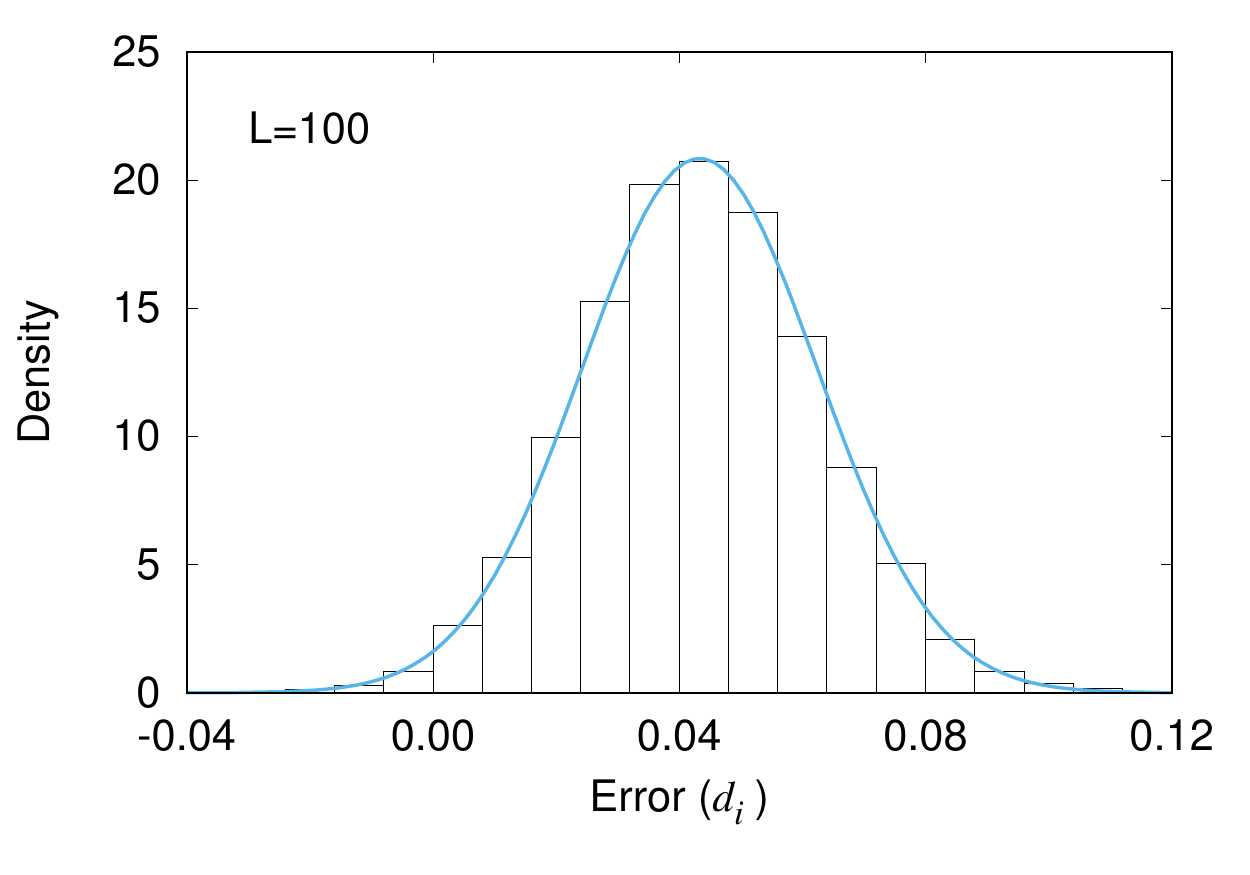}
  \includegraphics[width=0.49\textwidth]{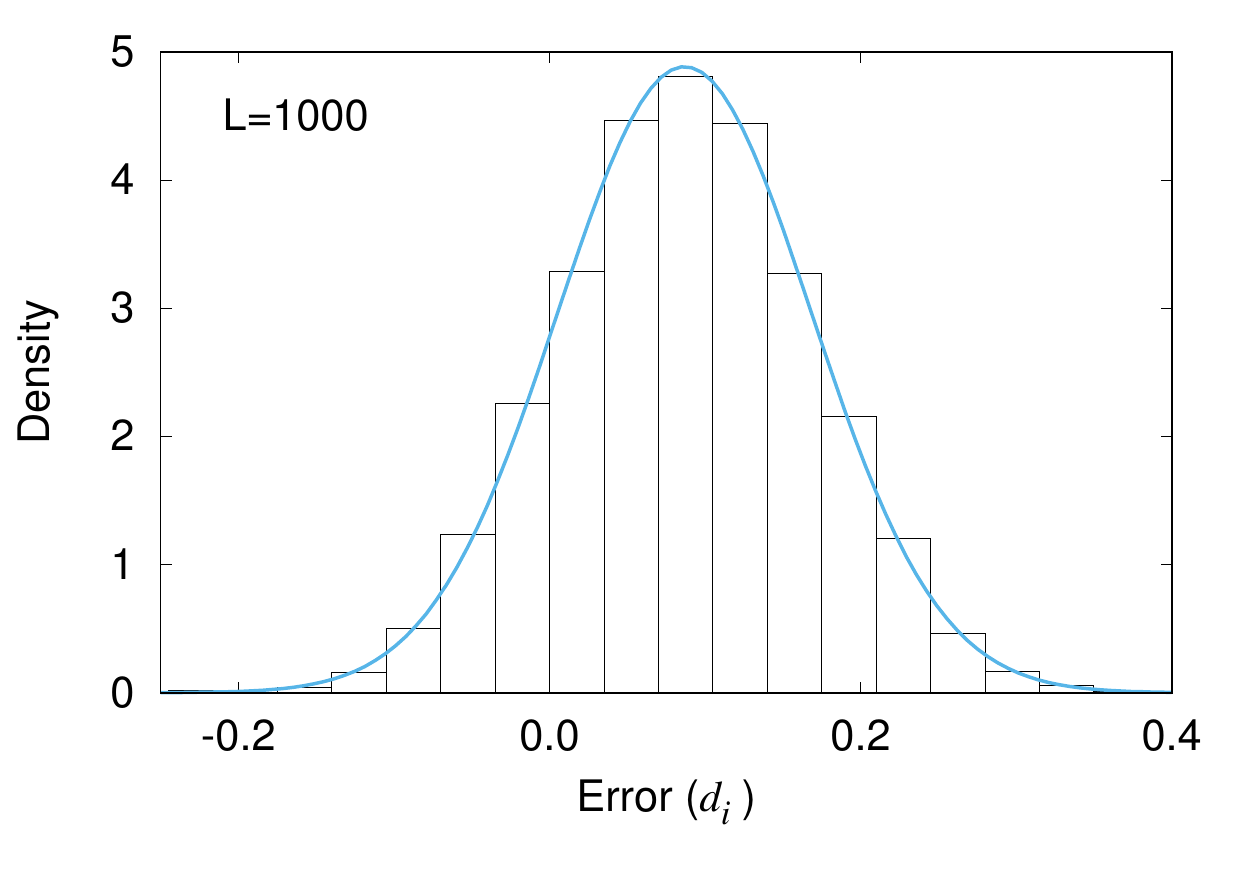}
  \vspace{-0.5cm}
  \caption{Distribution of the estimation error $d_i$, defined in Eq.~\protect\eqref{eq:est_err}, calculated with 10000 randomly generated $L=100$ (left) and $L=1000$ (right) matrices. Training parameters are given in Table~\protect\ref{tab:param}. Overlaid (blue) curve is the Normal distribution function with mean and standard deviation calculated from the distribution of $d_i$.\label{fig:err_hist}}
  \vspace{0.5cm}
  \includegraphics[width=0.49\textwidth]{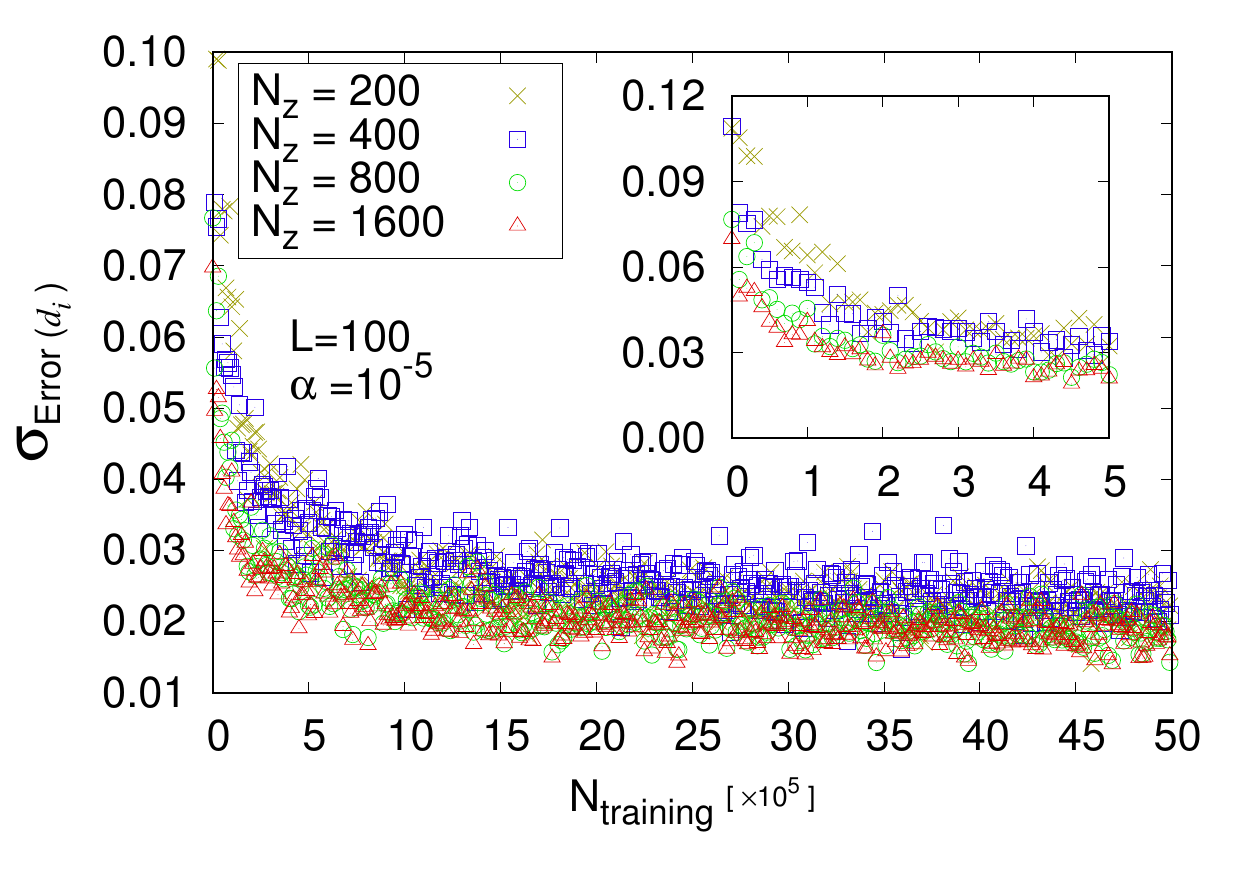}
  \includegraphics[width=0.49\textwidth]{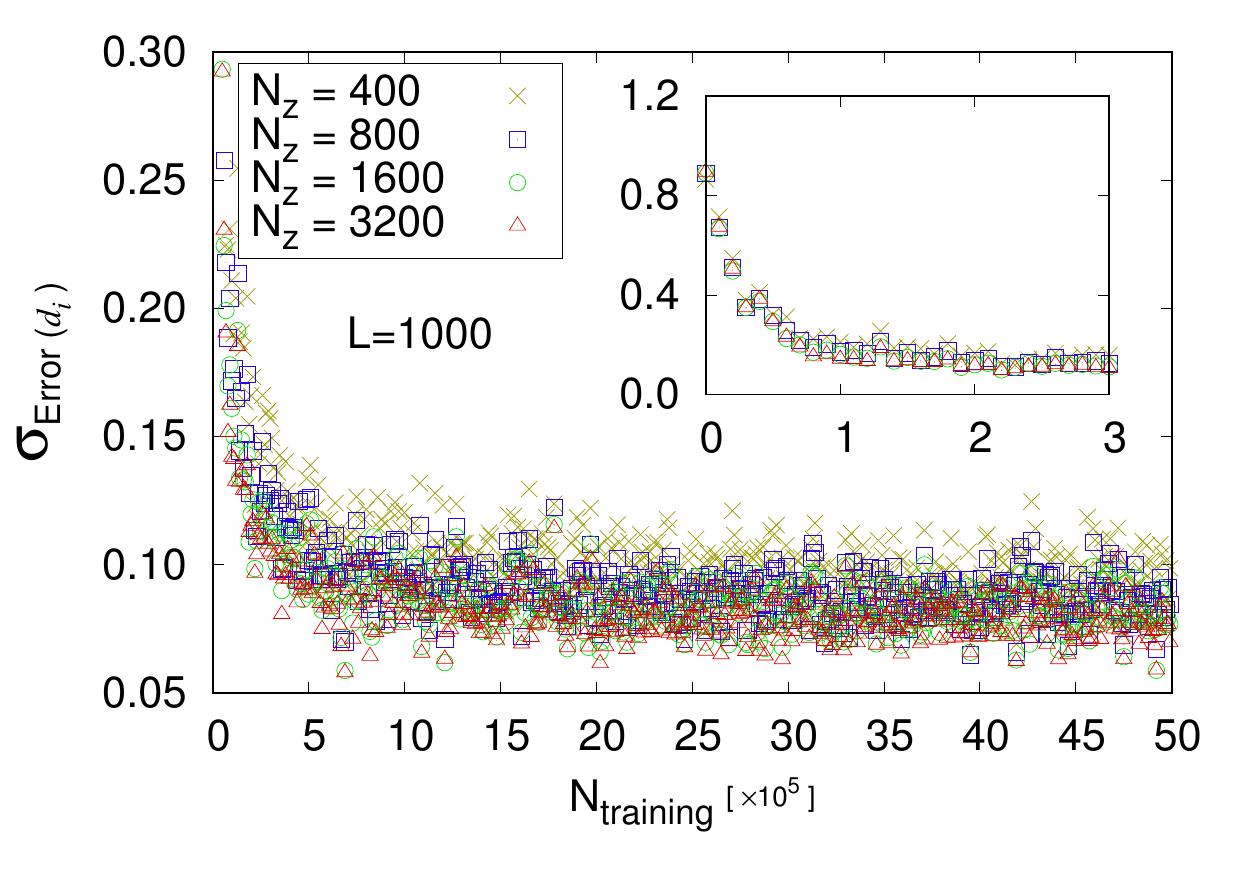}
  \vspace{-0.5cm}
  \caption{Standard deviation of the estimation error as a function of the number of input matrices in the training for different number of noise vectors of the Hutchison stochastic trace estimator used in the trace calculation of the input matrices. The training parameters given in Table~\protect\ref{tab:param} are used, except the number of noise vectors ($N_z$). The standard deviations are calculated from the $d_i$ results on randomly generated 50 test matrices.\label{fig:noise-dep}}
\end{figure}
\begin{figure}[tb]
\centering
  \includegraphics[width=0.49\textwidth]{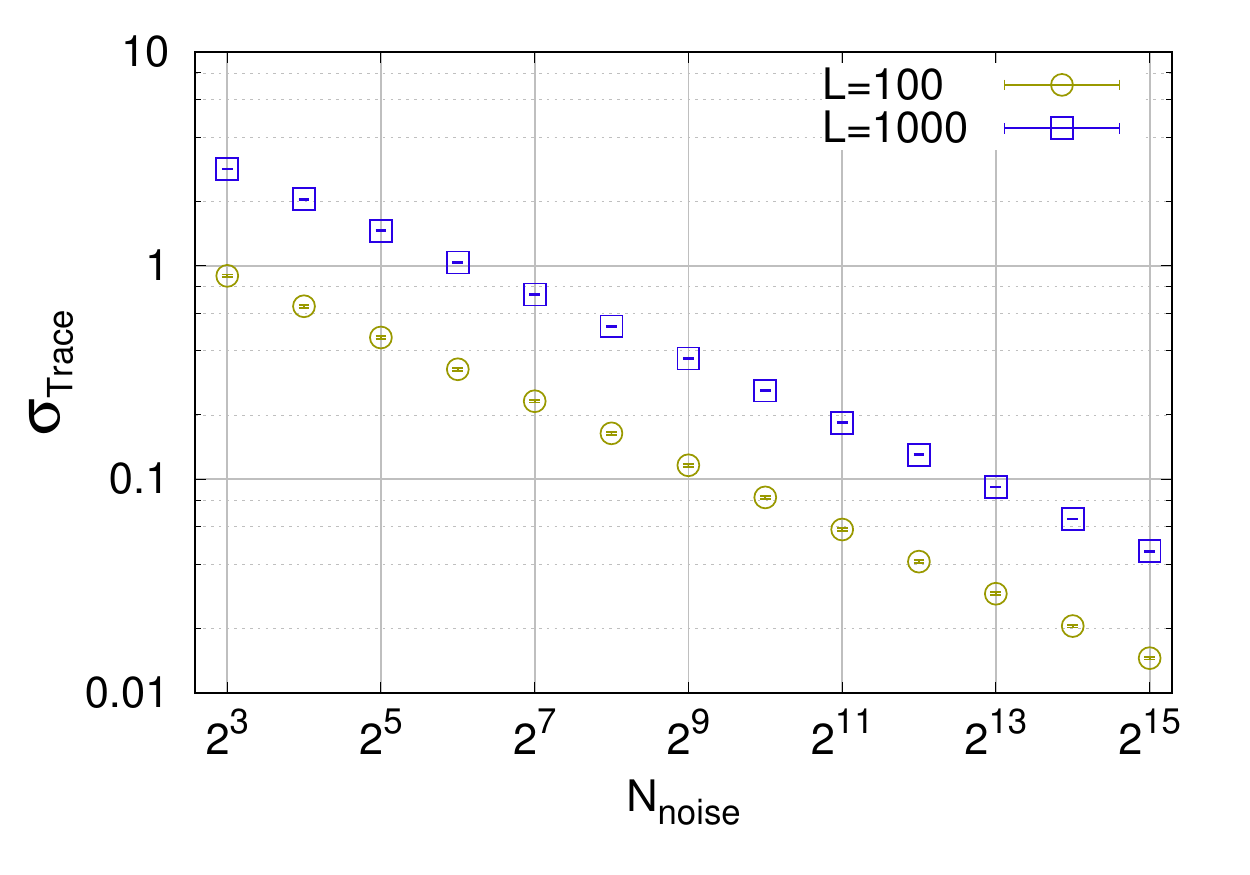}
  \vspace{-0.5cm}
  \caption{Uncertainty of the Hutchison stochastic trace estimator for different number of random noise vectors. The error bar includes both the uncertainty of the error of trace due to finite number of random noise vectors, and the fluctuation of the error of traces for random matrices generated following Eq.~\protect\eqref{eq:rand_mat}.\label{fig:tr-noise-dep}}
\end{figure}
\begin{table}
\centering
\begin{tabular}{c|cc|l}
\hline
L        & 100       & 1000      & Note \\ \hline
$N_p$    &   8       &   16      & Probing vectors \\
$N_z$    & 800       & 1600      & Random noise vectors \\
$\eta$   & 0.8       & 0.8       & Momentum term \\
$\alpha$ & $10^{-5}$ & $10^{-5}$ & Learning rate scheduling \\
$N_r$    & $10^5$    & $10^5$    & Pool size of input matrices \\
$N_\text{training}$ & $5 \times 10^6$ & $5 \times 10^6$ & Trainings \\
\hline
\end{tabular}
\caption{Training parameters for two different sizes of matrices tested in this paper.
\label{tab:param}}
\end{table}

We test two different sizes of matrices: $L=100$ and $1000$. Note that the training requires input matrices that have similar structures and the traces of each matrix. Although it is possible to use the exact trace calculations on those sizes of matrices we are testing in this section, we use estimates from the Hutchison estimator Eq.~\eqref{eq:tr_stoch} for the traces of the input matrices so that the results are scalable to larger size matrices, whose exact traces are very expensive to calculate. The traces of inverse matrices are 70.92 $\pm$ 0.07 and 709.20 $\pm$ 0.22 for $L=100$ and $1000$, respectively. The numbers after ``$\pm$'' notation are the standard deviation of the traces of random matrices. Unless explicitly noted, the discussion in this section is using the training parameters given in Table~\ref{tab:param}.

In the simulation, we find that the distribution of the estimation error $d_i$, defined in Eq.~\eqref{eq:est_err}, is following the normal distribution, as shown in Fig.~\ref{fig:err_hist}. Hence we use the standard deviation of $d_i$ as the uncertainty of the estimates. Fig.~\ref{fig:noise-dep} shows the learning efficiency for different number of noise vectors used in the Hutchison trace estimator. It shows that the learning efficiency is saturating when the number of noise vectors is larger than 800 (1600) for $L=100$ (1000). One interesting point is that the final uncertainty of the trace estimator is much smaller than the accuracy of the Hutchison stochastic estimator we used in the training. By comparing the uncertainty of the Hutchison stochastic trace estimator given in Fig.~\ref{fig:tr-noise-dep}, one finds that the error of the trace estimator using probing vectors are similar to that of Hutchison stochastic estimator with about 20000 noise vectors.

\begin{figure}[tb]
\centering
  \includegraphics[width=0.49\textwidth]{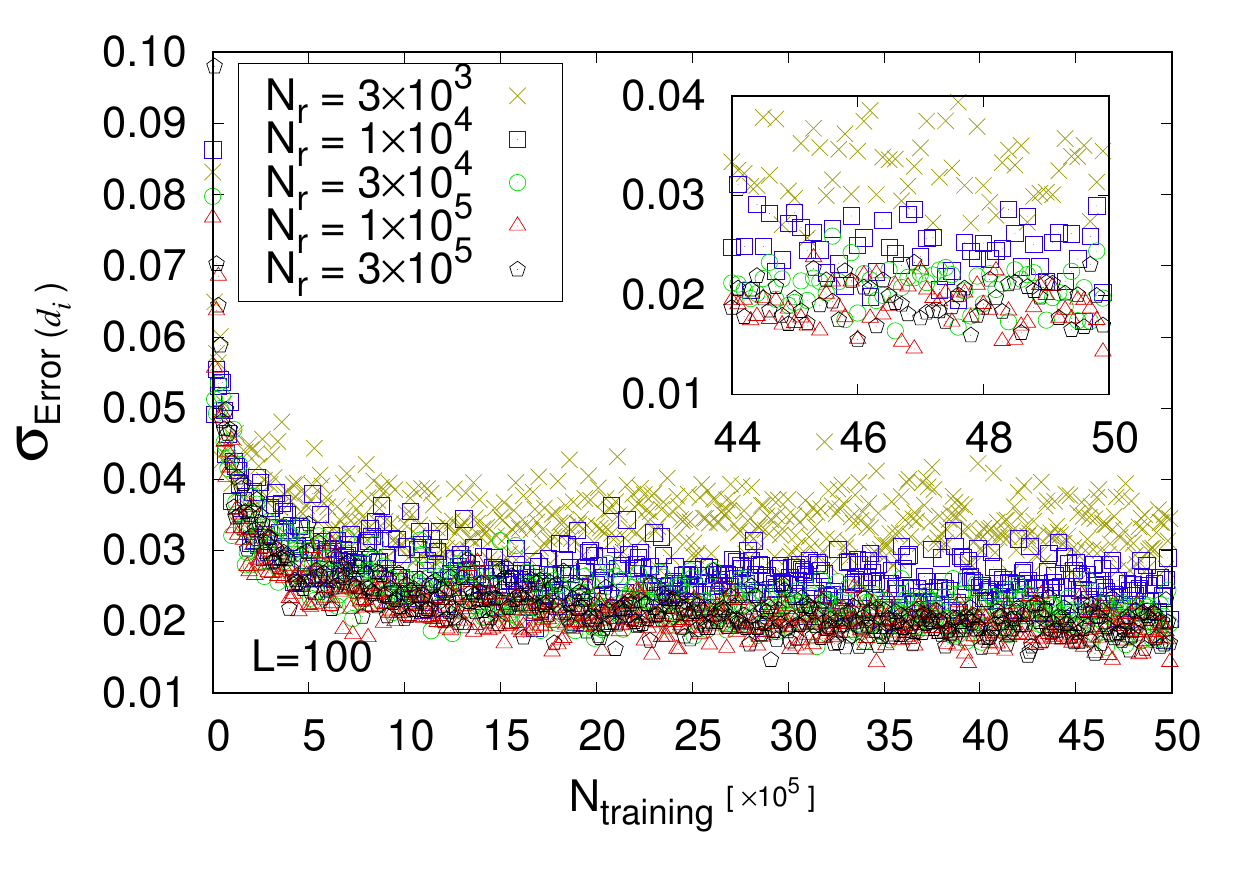}
  \includegraphics[width=0.49\textwidth]{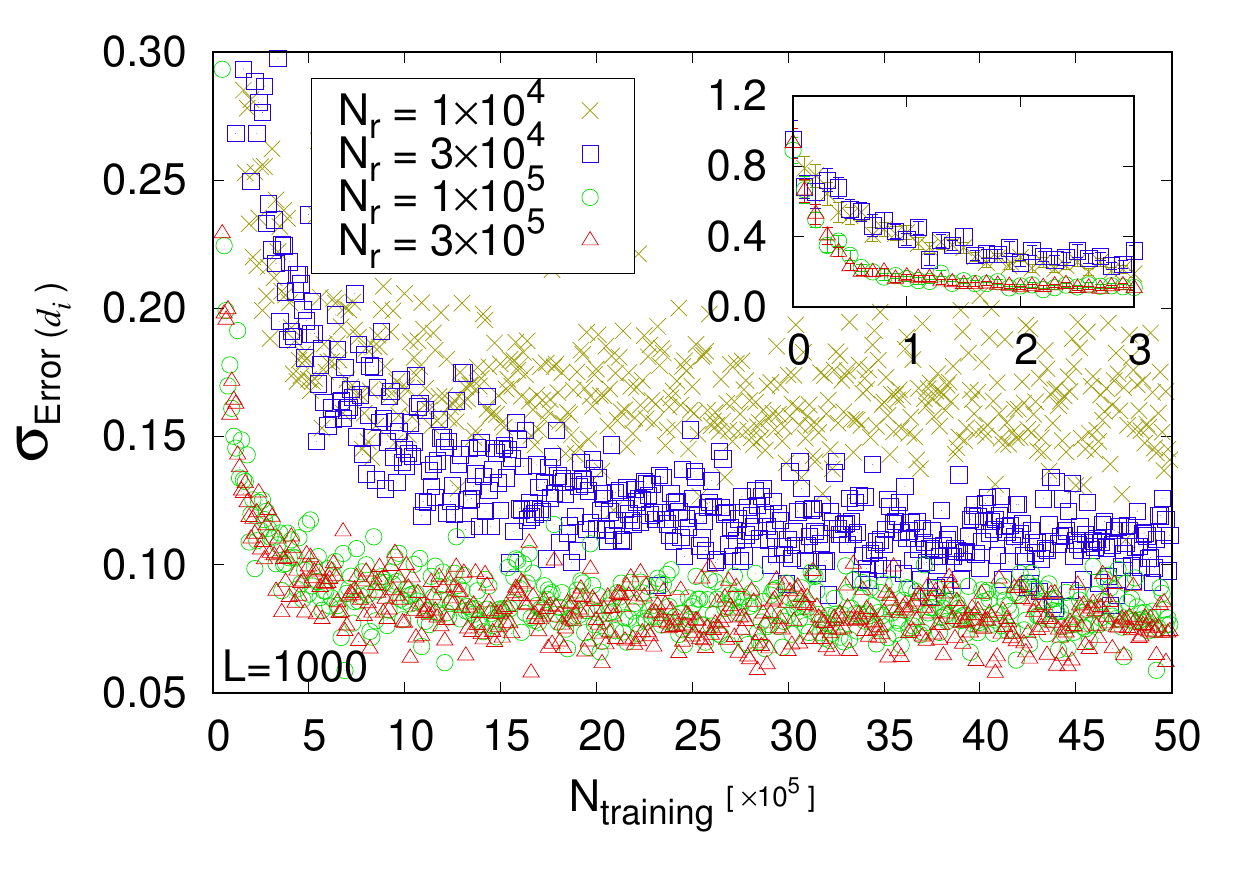}
  \vspace{-0.5cm}
  \caption{Standard deviation of the estimation error for different size of the pool of random matrices ($N_r$) for $L=100$ (left) and $L=1000$ (right). The input matrices for the training are randomly picked up from the pool, and the trace estimation of the inverse of the input matrix is done only once and reused when the same matrix is picked up repeatedly.\label{fig:reuse-dep}}
\end{figure}
\begin{figure}[tb]
\centering
  \includegraphics[width=0.49\textwidth]{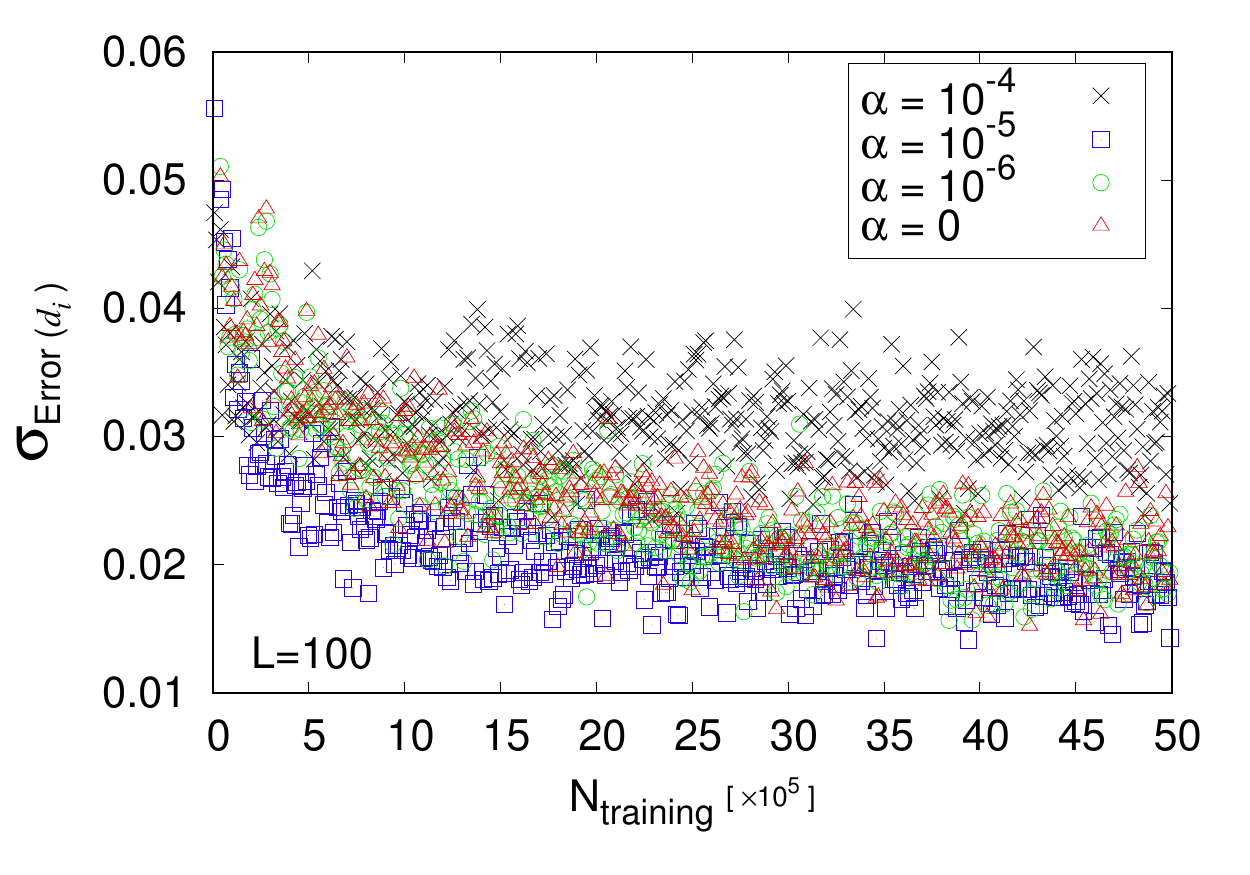}
  \includegraphics[width=0.49\textwidth]{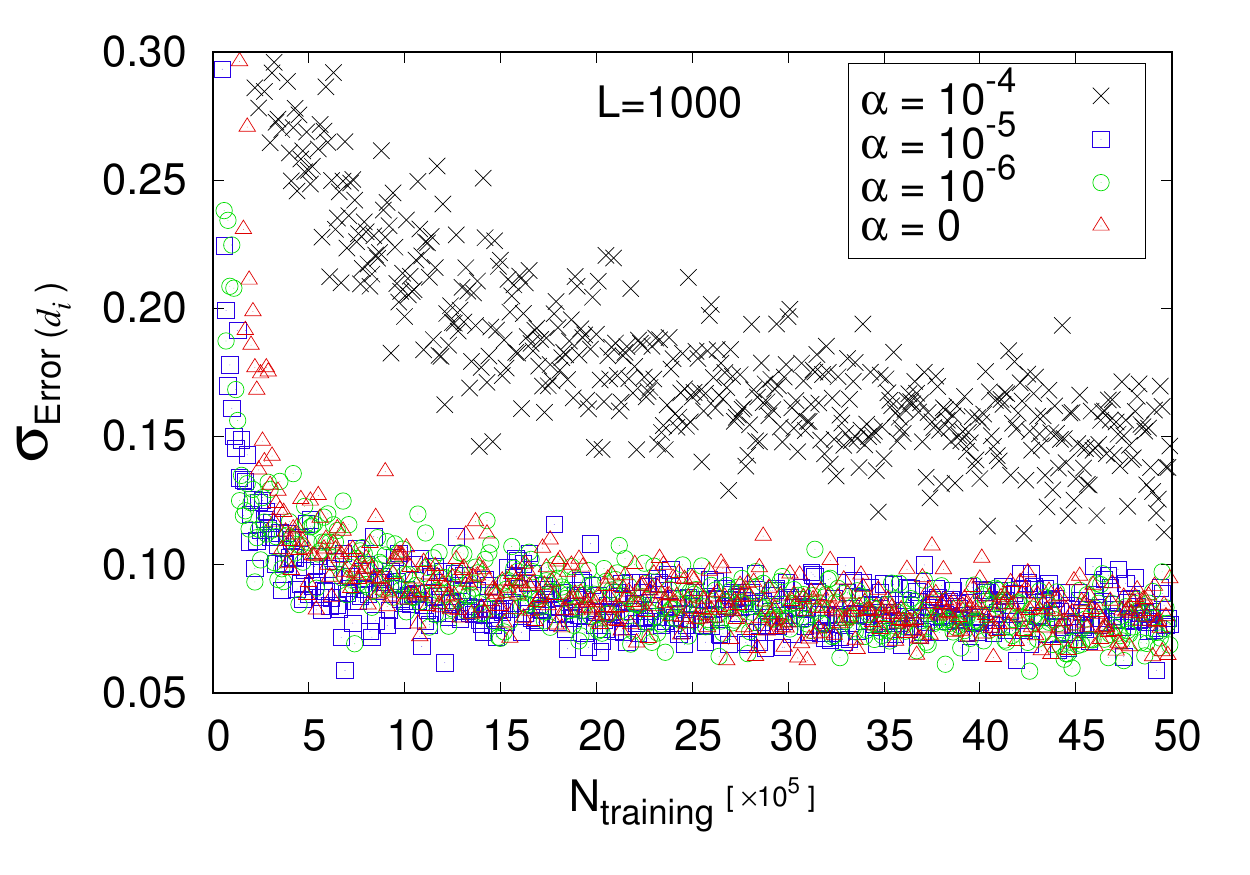}\\
  \includegraphics[width=0.49\textwidth]{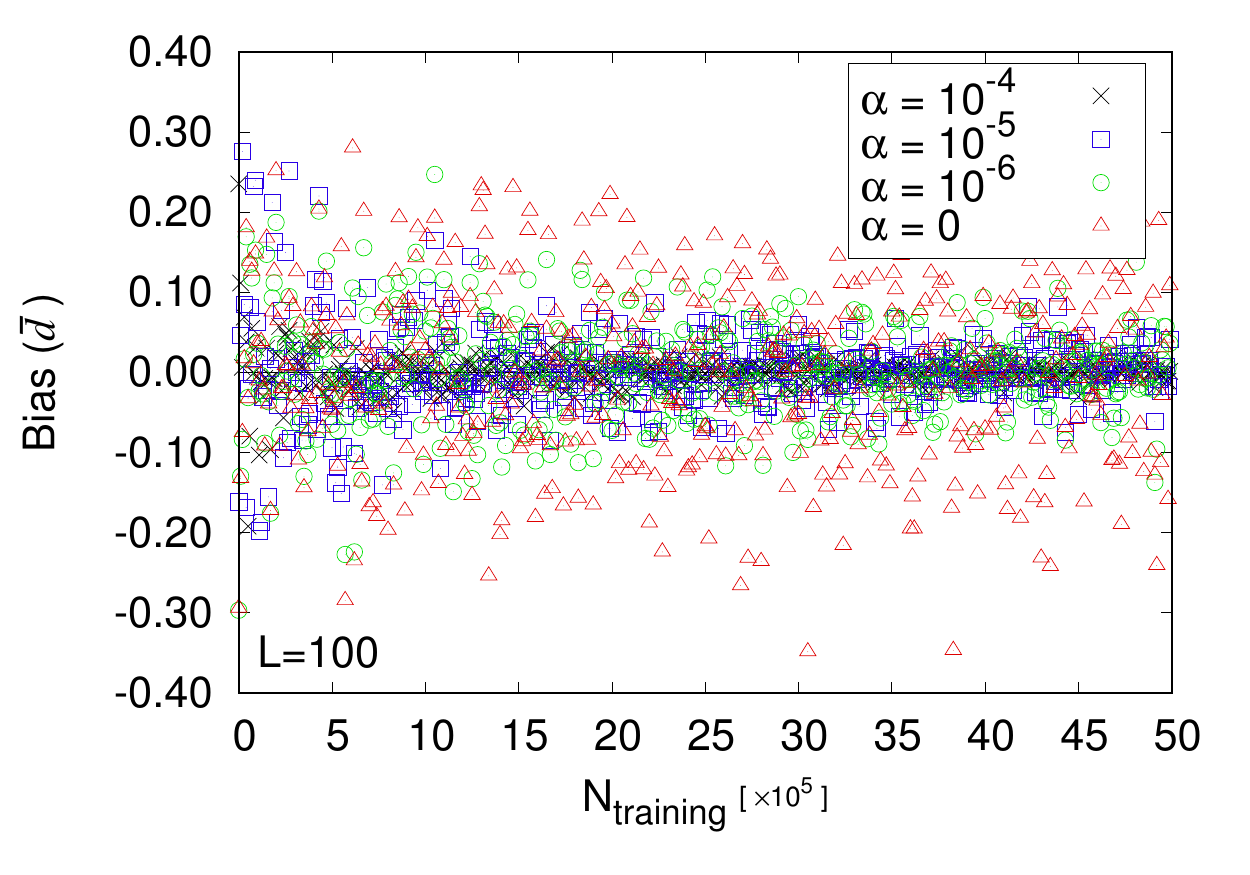}
  \includegraphics[width=0.49\textwidth]{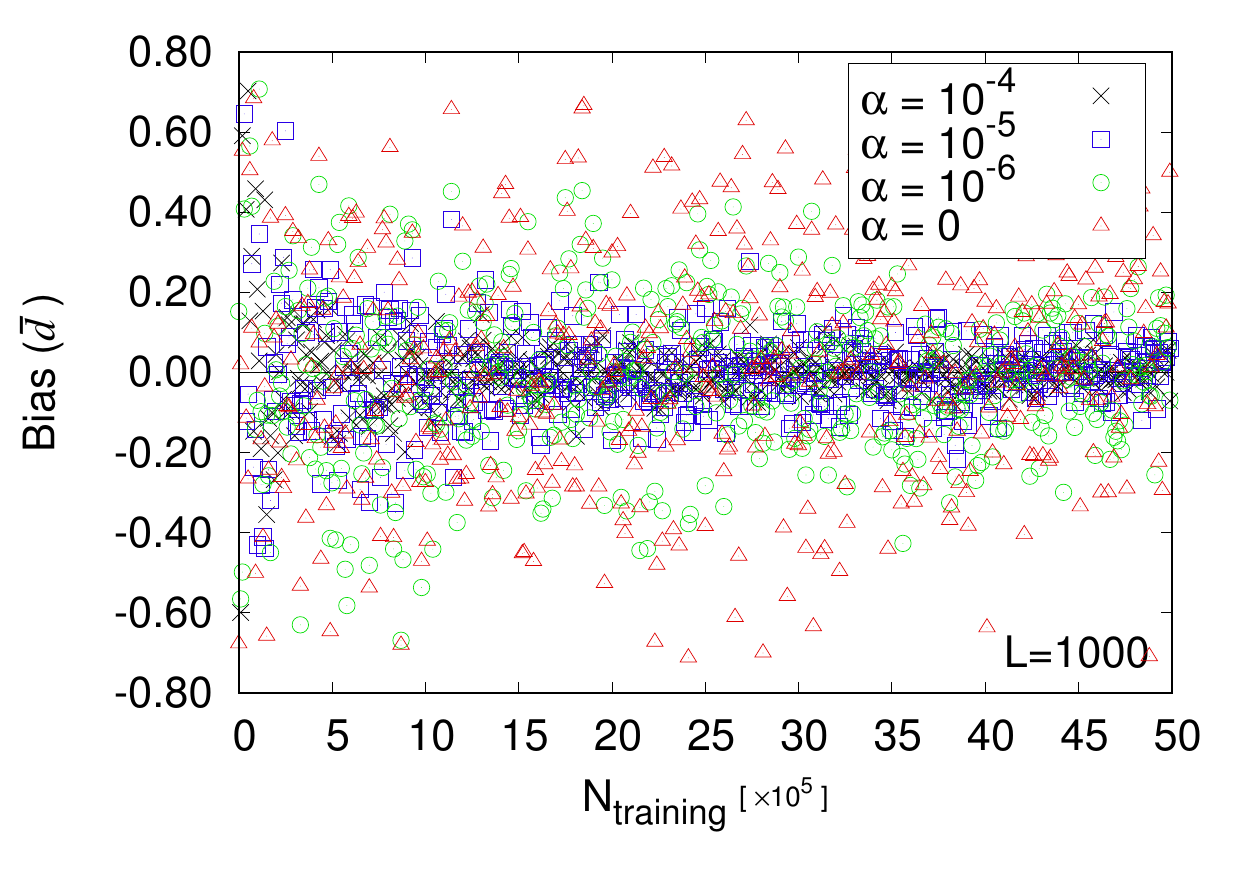}
  \vspace{-0.5cm}
  \caption{Standard deviation of the estimation error (upper) and average of the estimation error (lower) for different values of $\alpha$. The coefficient of the momentum term $\eta$ in Eq.~\protect\eqref{eq:w_update} is chosen differently for different values of $\alpha$ in order to maximize the learning efficiency: we use $\eta = 0.8$ for $\alpha = 10^{-4}$ and $10^{-5}$, and $\eta=0.5$ for $\alpha = 0$ in both $L=100$ and $1000$ cases, $\eta = 0.5$ for $\alpha = 10^{-6}$ in $L=100$, and $\eta = 0.65$ for $\alpha = 10^{-6}$ in $L=1000$ trainings. In general, smaller $\alpha$ requires larger $\eta$ values.\label{fig:alpha-dep}}
\end{figure}

Although we use relatively small number of noise vectors for the trace estimation of the input matrices for training, the most expensive part of the computation is the trace estimation. Hence we generate a finite size of pool of input matrices, and randomly pick up input matrices for the training from the pool. In the training procedure, the same matrix can be picked up multiple times. Reusing the same trace estimation calculated at the first pick-up significantly reduces the training cost. Fig.~\ref{fig:reuse-dep} shows the learning efficiency for different size of the input matrix pool. It shows that the pool size of $10^5$ input matrices is large enough to give reasonable training results.

\begin{figure}[tb]
\centering
  \includegraphics[width=0.49\textwidth]{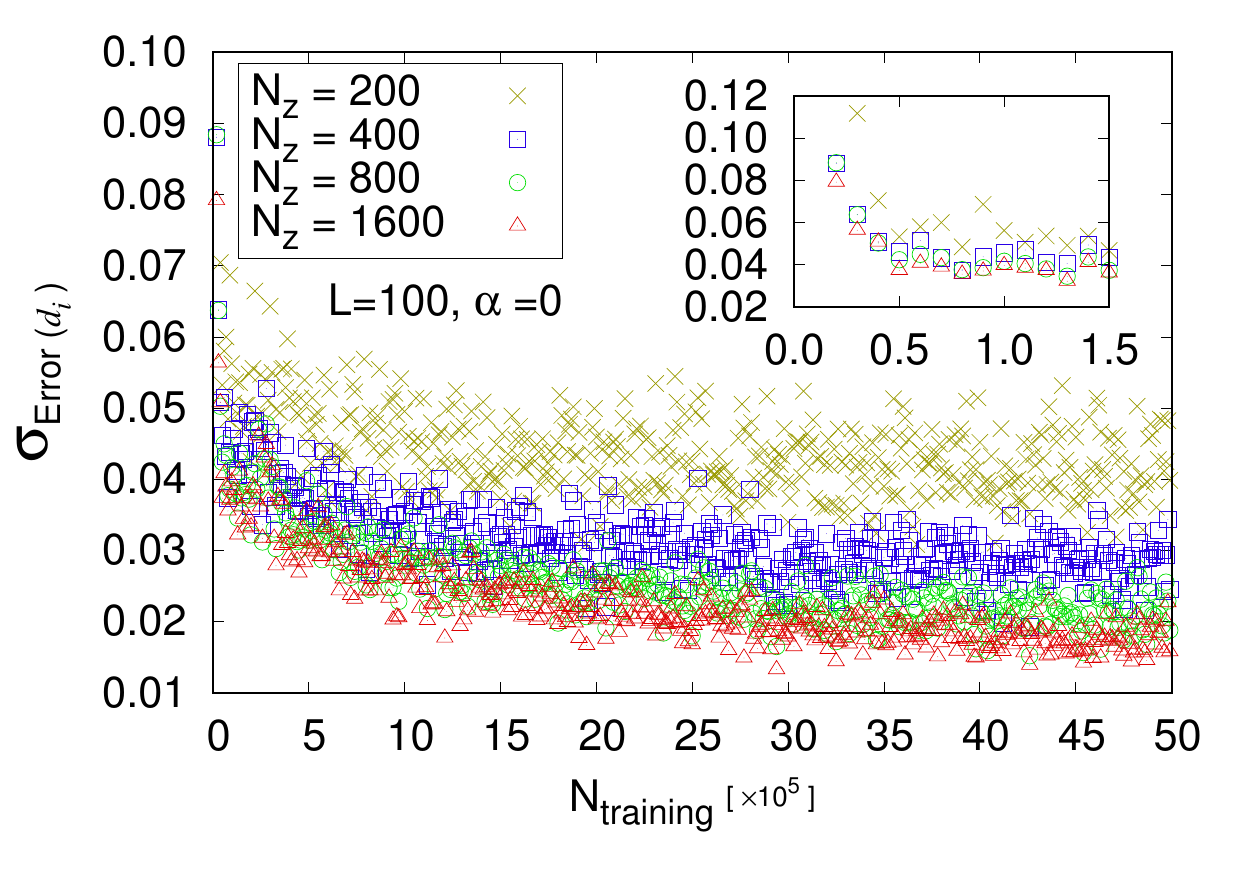}
  \vspace{-0.5cm}
  \caption{Standard deviation of the estimation error for $L=100$ matrices with $\alpha=0$. Notations are the same as those in Fig.~\protect\ref{fig:noise-dep}. \label{fig:noise-dep2}}
\end{figure}

Another important parameter in the training is the $\alpha$ that determines the scheduling of the learning rate, Eq.~\eqref{eq:learning_rate}. If the $\alpha$ is too large, the learning rate becomes tiny in early stage of the training, and it takes very long training period to converge. If the $\alpha$ is too small, the trace estimator using probing vectors tends to have large bias for finite training period. Fig.~\ref{fig:alpha-dep} shows the learning efficiency and bias of the estimator for different values of $\alpha$. As expected, the training is not efficient when $\alpha$ is too small or too large, but the bias is smaller for the trainings with smaller $\alpha$ values. 

As discussed in the previous section, the bias can be easily corrected by using the formulae given in Eq.~\eqref{eq:probe_est2} or Eq.~\eqref{eq:unbiased_est}, so the bias is not a concern in choosing a proper value of $\alpha$. However, the smaller value of $\alpha$ needs larger number of noise vectors for the trace estimation of training input matrices, as shown in Fig.~\ref{fig:noise-dep2}. The figure shows the training efficiency for different number of noise vectors when $\alpha=0$. By comparing it with Fig.~\ref{fig:noise-dep}, one finds that the training efficiency of $\alpha=0$ training is worse than that of $\alpha=10^{-5}$ training when the number of noise vectors is small. This is because the learning rate is relatively large when $\alpha$ is small. When the learning rate is large, training results are shifting for trainings of each input matrix. If the traces of input matrices are poorly determined, the training results are contaminated and fluctuating by the errors of the trace estimations of the input matrices. When learning rate is small, however, training results depend on series of input matrices, not only on the last one input matrix, so the errors of trace estimation of the input matrices are averaged out.

\begin{figure}[tb]
\centering
  \includegraphics[width=0.49\textwidth]{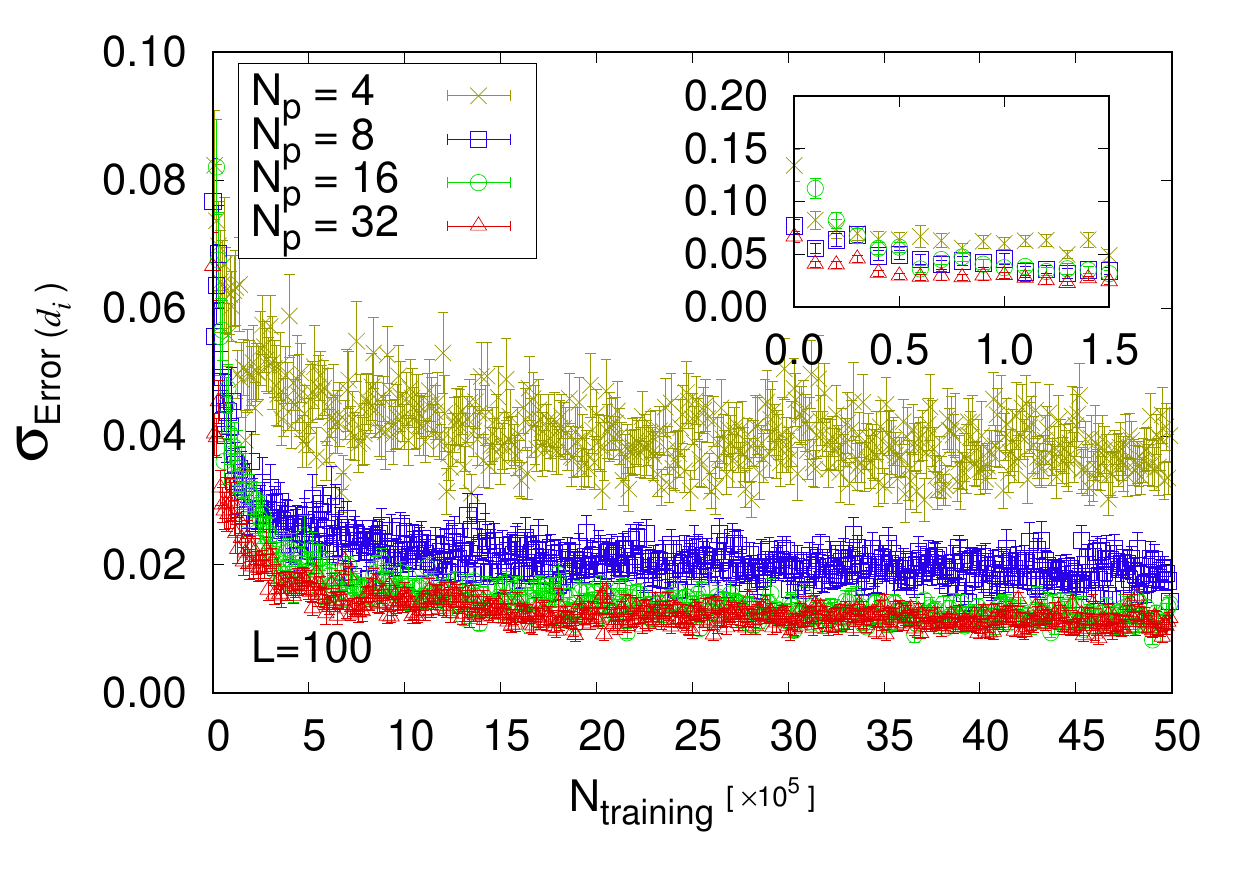}
  \includegraphics[width=0.49\textwidth]{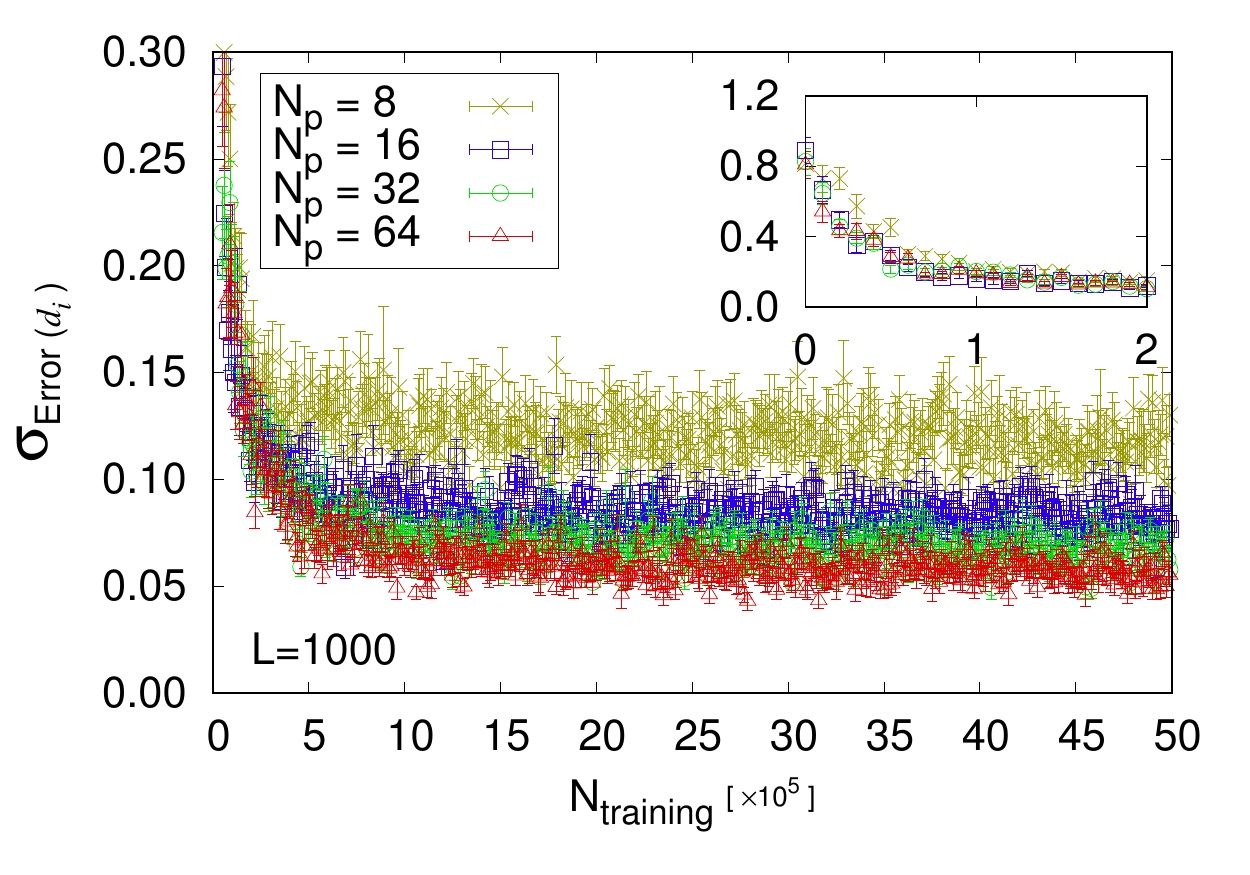}
  \vspace{-0.5cm}
  \caption{Standard deviation of the estimation error for different number of probing vectors $N_p$. Other training parameters are given in Table~\protect\ref{tab:param}. Error bars are the statistical uncertainty of the standard deviation calculated from randomly generated 50 test matrices. \label{fig:probe-dep}}
\end{figure}

Fig.~\ref{fig:probe-dep} shows the training efficiency for different number of probing vectors $N_p$. As expected, the more probing vectors give the smaller estimate uncertainty. Except the smallest number of probing vector cases, the estimation errors after the training are about $0.01\sim 0.02$ for $L=100$, and $0.05\sim 0.10$ for $L=1000$, depending on the number of probing vectors. By comparing the results with the efficiency of the Hutchison stochastic estimator, Fig.~\ref{fig:tr-noise-dep}, one finds that the accuracy of the probing vector trace estimation with about 30 probing vectors is similar to that of the Hutchison estimator with $20000\sim 30000$ random noise vectors.

\section{Conclusion}
\label{sec:sum}
We proposed a new trace estimator of the matrix whose explicit form is difficult to obtain but its matrix multiplication to a vector is easy to evaluate. The form of the estimator is similar to the Hutchison stochastic trace estimator, but instead of the random noise vectors in Hutchison estimator, we use small number of probing vectors determined by the machine learning technique. Through the training procedure, the probing vectors become probes for the trace of a matrix with given structure. Similar to other machine learning applications, the training procedure is computationally expensive, as it requires unbiased trace estimations for the input matrices, and multiple matrix-vector multiplications. Once training is over, however, it becomes an efficient trace estimator that requires only small number of matrix-vector multiplications. In the numerical experiments, it is shown that the accuracy of the trace estimator with $\mathcal{O}(10)$ probing vectors is similar to that of the Hutchison estimator with $\mathcal{O}(10000)$ random noise vectors. Comparing at the same number of vectors, the estimator with probing vectors has more than 20 times smaller uncertainty than the Hutchison estimator with random noise vectors. The error of the probing vector estimator can be further reduced if one extends the training period with retuned parameters.

The quality of estimates are evaluated by using the deviation of the estimate from the exact trace, $d_i$, defined in Eq.~\eqref{eq:est_err}. In the numerical experiments we performed, it turns out that $d_i$ is following the normal distribution. Hence we use the standard deviation of $d_i$ as an uncertainty of the trace estimates. The trace estimates of the probing vector estimators can be biased, but a bias correction method using $d_i$ is introduced in Eq.~\eqref{eq:probe_est2}. Another unbiased estimator, which converts the systematic error of the trace estimator to a statistical uncertainty, is presented in Eq.~\eqref{eq:unbiased_est}.

In this paper, we explored a general trace estimator using the machine learning idea. One might be able to define a better trace estimator by exploiting the symmetry conditions in a given system. Furthermore, the idea of exploring a matrix structure by using machine learning would be applicable to many other matrix problems.

%

\section*{Acknowledgements}
We thank Garrett Kenyon for discussions on machine learning algorithms. This research used resources of the Institutional Computing at Los Alamos National Laboratory, and the Oak Ridge Leadership Computing Facility at the Oak Ridge National Laboratory, which is supported by the Office of Science of the U.S. Department of Energy under Contract No. DE-AC05-00OR22725. This work is supported by the U.S. Department of Energy, Office of Science, Office of High Energy Physics under Contract No. DE-KA-1401020, and the LANL LDRD program.

\clearpage
%
\bibliographystyle{unsrt} 
\bibliography{ref} 

\end{document}